%% file: main.tex
\documentclass[lettersize,journal]{IEEEtran}
\usepackage{amsmath,amsfonts}
\usepackage{algorithmic}
\usepackage{algorithm}
\usepackage{array}
\usepackage[caption=false,font=normalsize,labelfont=sf,textfont=sf]{subfig}
\usepackage{textcomp}
\usepackage{stfloats}
\usepackage{url}
\usepackage{verbatim}
\usepackage{graphicx}
\usepackage{cite}

\usepackage{booktabs}
\usepackage{multirow}
\usepackage{makecell}
\usepackage{longtable}
\usepackage{xcolor}
\begin{document}

\title{
Enabling Weakly-Supervised Temporal Action Localization from On-Device Learning of the Video Stream}

\author{Yue Tang, Yawen Wu, Peipei Zhou and Jingtong Hu~\IEEEmembership{
}
\thanks{Yue Tang, Yawen Wu, Peipei Zhou and Jingtong Hu are with the Department of Electrical and Computer Engineering, University of Pittsburgh, Pittsburgh, PA 15261 USA.}
\thanks{Manuscript received April 07, 2022; revised June 11, 2022; accepted July 05, 2022. This article was presented in the International Conference on 2022 and appears as part of the ESWEEK-TCAD special issue.}}




\maketitle
\input{0_abstract}
\input{1_intro}

\input{2_related}
\input{3_motivation}
\input{4_algorithm}

\input{5_strategy}
\input{6_exp}
\input{7_conclusion}

\section*{Acknowledgments}
This work is supported by NSF CNS-2122320 and the University of Pittsburgh New Faculty Start-up Grant.



 
%

\bibliographystyle{IEEEtran}
\bibliography{wta}



 




\vfill

\end{document}

%% file: 0_abstract.tex
\begin{abstract}
Detecting actions in videos have been widely applied in on-device applications such as cars, robots, etc. Practical on-device videos are always untrimmed with both action and background. It is desirable for a model to both recognize the class of action and localize the temporal position where the action happens. Such a task is called temporal action location (TAL), which is always trained on the cloud where multiple untrimmed videos are collected and labeled. It is desirable for a TAL model to continuously and locally learn from new data, which can directly improve the action detection precision while protecting customers’ privacy. However, directly training a TAL model on the device is non-trivial. To train a TAL model which can precisely recognize and localize each action, tremendous video samples with temporal annotations are required. However, annotating videos frame by frame is exorbitantly time-consuming and expensive. Although weakly-supervised temporal action localization (W-TAL) has been proposed to learn from untrimmed videos with only video-level labels, such an approach is also not suitable for on-device learning scenarios. In practical on-device learning applications, data are collected in streaming. For example, the camera on the device keeps collecting video frames for hours or days, and the actions of nearly all classes are included in a single long video stream. Dividing such a long video stream into multiple video segments requires lots of human effort, which hinders the exploration of applying the TAL tasks to realistic on-device learning applications. To enable W-TAL models to learn from a long, untrimmed streaming video, we propose an efficient video learning approach that can directly adapt to new environments. We first propose a self-adaptive video dividing approach with a contrast score-based segment merging approach to convert the video stream into multiple segments. Then, we explore different sampling strategies on the TAL tasks to request as few labels as possible. To the best of our knowledge, we are the first attempt to directly learn from the on-device, long video stream. Experimental results on the THUMOS’14 dataset show that the performance of our approach is comparable to the current W-TAL state-of-the-art (SOTA) work without any laborious manual video splitting. 
\end{abstract}

\begin{IEEEkeywords}
weakly-supervised temporal action localization (W-TAL), on-device learning, video understanding task, raw video stream,
video splitting free algorithm.
\end{IEEEkeywords}

%% file: 1_intro.tex
\section{Introduction}
\label{sec:Introduction}
{D}{etecting} actions in videos have been widely applied in lots of on-device applications such as autonomous driving~\cite{wang2018semi}, medical care~\cite{jin2020diagnosing}, intelligent surveillance~\cite{sreenu2019intelligent}, etc. 
Practical on-device video is captured as a continuous stream from cameras. 
The video is always untrimmed. 
Untrimmed means the video has both multiple actions and backgrounds. For on-device video detection, it is desirable for a model to both recognize the class of action and localize the temporal position where the action happens. For example, if a medical care robot detects an elderly falling down, reporting the action of falling down and bounding box the frames where the action happens can help rescuers analyze the patient's situation and provide the most appropriate cure. Such a video detection task is called temporal action location (TAL) and has attracted lots of researchers' attention~\cite{shou2016temporal,chao2018rethinking}.

\IEEEpubidadjcol

Although the TAL task is a practical on-device video detection task, such a task is always trained on the centralized cloud where multiple untrimmed videos are collected and labeled. 
However, when the device comes to a new environment, the model needs to be updated for domain adaption. The process of transmitting local video frames to the cloud, waiting for the center to retrain the model, and receiving the new model is inefficient and time-consuming. 
Besides, on-device learning can provide better privacy since users do not need to upload data into the centralized cloud~\cite{xu2018deeptype}. Therefore, it is desirable for a TAL model to continuously and directly learn from local data on the device. 
However, directly training a TAL model on the device is non-trivial. To train a TAL model which can precisely localize and recognize each action, tremendous data are required with temporal annotations. Generating such annotations requires humans to carefully watch the videos frame by frame, which is exorbitantly time-consuming. 
Besides, there is no explicit boundary between actions and background which makes it difficult to manually provide temporal annotations, and providing such temporal annotations is prone to cause errors~\cite{gong2020learning}. 
Therefore, weakly-supervised temporal action localization (W-TAL)~\cite{paul2018w,zhang2021cola,su2021improving} has been proposed which can learn from untrimmed videos with only video-level labels. Although it can save lots of energy and cost, such an approach is still not suitable for on-device scenarios because of the following issues.

Firstly, current W-TAL works~\cite{paul2018w,zhang2021cola,su2021improving} consider each video as a learning sample with a video-level label, and each video only includes actions from limited classes. For example, most videos contain actions only from one class in THUMOS’14~\cite{idrees2017thumos} and ActivityNet~\cite{caba2015activitynet} datasets which are commonly used in TAL tasks. 
Only a few videos contain actions from 2 or 3 classes. However, in practical on-device applications, data are collected in streaming. The camera on the device keeps collecting video frames in hours or days, and the actions of nearly all classes are included in a single long video. 
Separating the video into individual video segments and providing video-level labels also requires lots of human labeling costs. If a divided segment is too short, it cannot contain complete action and background information which 
can lead to low learning performance, and more video-level labels are required under the same length of a stream. If the segment is too long, it will contain actions from multiple classes which increases the TAL learning difficulty. Besides, longer segments mean fewer learning samples under the same length of the streaming video, while the lack of samples may also lead to inferior learning performance. 
Second, in order to provide a video-level label for different video segments, the whole video needs to be uploaded to the oracle of the cloud-level data center. 
The whole video stream which lasts hours or days takes up tens to hundreds of GB spaces. Transmitting such a huge amount of data from local devices to the data center is inefficient.

To tackle the challenges and enable weakly-supervised temporal action localization learning from a long on-device video stream without laborious manual video splitting, we propose a self-adaptive video dividing approach with a contrast score-based segment merging approach that divides the long video stream
into individual video segments during the training process. 
We also explore different sampling strategies to request as few labels as possible. 

Our main contributions are as follows.

\begin{itemize}
\item  We propose an efficient video learning approach that can directly adapt to new environments from a single on-device long video stream without laborious manual video splitting. To the best of our knowledge, we are the first attempt to explore TAL tasks in such a streaming learning case rather than training the TAL model from separate short untrimmed videos compared to state-of-the-art (SOTA) works. 
The overview of our streaming learning method will be introduced in Section~\ref{sec:motivation}. 

\item Based on our streaming learning workflow, we propose a self-adaptive video dividing approach with a contrast score-based segment merging approach which splits the continuous video stream into individual video segments during the training process. 
We first evenly divide the stream into segments when the stream is collected and then apply the contrast score-based approach to decide whether two adjacent segments need to be merged or not to convey complete action and background information. The approach is self-adaptive in the training process, which will be explained in Section~\ref{sec:segment}.

\item 
We propose an interests-based sampling strategy that selects the segments that contain more interesting areas, i.e. the areas that are more possible to contain actions. The discussion will be explained in Section~\ref{sec:sample}.
\end{itemize}

%% file: 2_related.tex
\section{Related Works}
\label{sec:Related works}

\subsection{Temporal Action Localization}
\label{sec:TAL}
In previous video learning works~\cite{tran2018closer,feichtenhofer2020x3d}, a video is extracted into multiple frames with a certain frames per second (fps), and a given number of contiguous frames compose a clip. 
For a trimmed clip of video that only contains a single action, a deep neural network (DNN) model is developed to recognize the action of the video.
The TAL~\cite{shou2016temporal,chao2018rethinking} task has been proposed to work on untrimmed videos and has attracted lots of researchers' attention.

Given an untrimmed video, TAL mainly solves
two tasks: 1) when does the action occur, i.e., the start time and the
end time of the action; 2) what category does each action proposal belong to~\cite{xia2020survey}. 
Current SOTA works learn their TAL models from a large number of untrimmed videos where each video contains  both actions and backgrounds. 
The video is extracted into multiple frames with a given fps (for example 25fps is commonly used to extract a video into frames), and a given number of contiguous frames compose a clip (e.g. 16 frames per clip, 64 frames per clip, etc.). 
In 2016, S-CNN~\cite{shou2016temporal} proposed a multi-stage framework including a proposal network to identify candidate segments in an untrimmed video that may contain actions, a classification network learning one-vs-all action classification model to serve as initialization for the localization network, and a localization network fine-tuning the learned classification network to
localize each action instance. AFSD~\cite{lin2021learning} proposed an anchor-free TAL framework generating only
one proposal for each temporal location without adjusting the pre-defined anchors. 
Although these methods can precisely detect actions in untrimmed videos, their TAL models need to be learned from full labeled videos with both video-level labels and temporal annotations. 
However, generating such annotations requires humans to carefully watch the videos frame by frame, which is exorbitantly time-consuming and prone to cause errors.

To reduce the human efforts in providing temporal annotations, weakly-supervised TAL has been proposed to learn from untrimmed videos with only video-level labels. 
W-TALC~\cite{paul2018w} proposed a framework with sub-networks: the Two-Stream-based feature extractor network and a weakly-supervised module. The feature extractor used the I3D~\cite{carreira2017quo} network pre-trained on the Kinetics dataset~\cite{carreira2017quo} as the backbones to extract RGB and optical features. 
The weakly-supervised module took the frozen features as the input and got improved by minimizing the Multiple Instance Learning Loss (MILL) and the Co-Activity Similarity Loss (CASL). 
Contrastive learning (CL)~\cite{chen2020simple,he2020momentum,wu2020enabling} further advances W-TAL. 
CL learns representations from unsupervised data by minimizing noise contrastive estimation (NCE)~\cite{gutmann2010noise} loss between negative pairs and positive pairs, i.e. minimizing the distance of features from associated data and maximizing the distance of features from distinctive data, 
By adopting CL, W-TAL works began to distinguish the boundary of action and background without frame-level annotations. 
For example, CoLA~\cite{zhang2021cola} is the first to introduce noise contrastive estimation to W-TAL tasks. 
It proposed a Hard \& Easy Snippet Mining strategy to select clips of easy action (EA), easy background (EB), hard action (HA), and hard background (HB) and proposed a Snippet Contrast (SniCo) loss to enable the embedded features of HA closer to that of EA and the embedded features of HB closer to that of EB. 
Although the W-TAL works have improved a lot without requesting temporal annotations, it is still difficult to imply W-TAL in on-device applications. 
In on-device learning scenarios, the camera keeps collecting video frames for hours or days. Unlike previous works that a TAL model is learned from separate videos with actions of limited categories, such a long video stream collected from the local camera may contain actions of many categories. 
Separating the video into individual video segments and providing video-level labels also requires lots of human costs. 
In addition, in order to provide video-level labels, all the videos need to be first uploaded to the cloud data center to enable access from the oracle, i.e., human annotators.
The upload process takes up lots of transmission time and storage space.

To reduce the cost and effort in annotation, several works have been proposed to learn the TAL model from both labeled and unlabeled data. 
For example, Ji et. al.~\cite{ji2019learning} proposed a semi-supervised learning algorithm that outperformed the fully-supervised baseline with only 60\% of the videos fully labeled. 
Shi et. al.~\cite{shi2021temporal} introduced a semi-supervised action detection task with a mixture of fully labeled, weakly labeled, and unlabeled data. 
Gong et. al.~\cite{gong2020learning} proposed a two-step ``clustering + localization'' iterative procedure and became the first attempt to explore the TAL task under an unsupervised setting. 
However, this approach requires mapping each cluster to the action class in the testing step. 
Besides, same with W-TAL tasks, all of the above-mentioned works are learned from separate untrimmed videos and fail to consider the realistic on-device scenarios where the data is collected in a single long stream video without laborious manual splitting. 
\vspace{-6pt}

\subsection{Streaming Learning}
\label{sec:streaming}
Different from conventional training in the cloud data center, it is desirable for local devices to continuously learn from the newly generated streaming data. 
The tremendous input images are generated frame-by-frame with high frequency, and local devices do not have enough data storage space to store all the images.
A desirable way is to streamline the captured video and learn from the single pass of the streamline. 
Therefore, streaming learning can not use the conventional DNNs for two reasons: 1) conventional DNNs require large storage space to store the entire dataset, and multiple epochs of back propagation and feed forward are performed on the entire dataset, and 2) non-iid (independent and identically distributed) data will cause catastrophic forgetting~\cite{hayes2019memory}. 
To solve these problems, Hayes et. al.~\cite{hayes2019memory} explored different memory rehearsal approaches which sampled the incoming data and mixed the new samples with buffered samples. However, this work is based on supervised learning.
If deployed in realistic scenarios, supervised learning means acquiring labels for the local devices from the cloud oracle. This will incur a long latency in the transmission and usually can not meet the real-time requirement. 
To learn from the unlabeled data stream, Wu et. al.~\cite{wu2021enabling} applied CL for representation learning and proposed a framework with contrast scoring to automatically select the most
representative data from the unlabeled input stream. However, what they solve is to apply streaming learning for image classification tasks, i.e., non-iid problems, and they fail to discuss how to apply streaming learning on TAL tasks. 

To the best of our knowledge, none of the TAL works have considered learning from the streaming data. 
Different from image classification where the whole DNN model is updated after new data is coming, most TAL designs~\cite{paul2018w,zhang2021cola,gong2020learning} have frozen the encoder of the 3D-DNN and directly used the extracted RGB features and optical flow features as inputs. 
Therefore, it is feasible for local devices to store all the extracted features which have been collected for hours or days. 
However, unlike the image classification task, the main challenge is not on how to deal with the non-iid problem, but how to identify an individual learning sample. 
In image classification, each frame is a learning example that can be simply identified. 
For a long video stream, there is no explicit methodology on how to divide the video stream into separate video segments that can benefit the training performance. 
Manually dividing the video stream which lasts hours or days requires laborious human efforts, and there is also no guarantee that manual splitting can benefit the learning performance. Besides, when data comes in a streaming fashion, only the extracted features and important clips are collected. 
It is not desired that dedicated experts from the cloud oracle keep observing the input stream and split the stream into individual videos in real-time when the on-device system is operating. 
Therefore, a new approach is required to directly learn from the on-device, long video stream without laborious manual video splitting for the TAL task.
\vspace{-6pt}

\subsection{Active Learning}
\label{sec:AL}
Apart from dividing the video stream into individual video segments, it is also challenging to sample the most representative segments to request labels to release the labeling labor and transmission requirement. 
Active learning (AL)~\cite{ren2021survey} is a method aiming to select the most useful samples from the unlabeled dataset and hand it over to the oracle for
labeling to reduce the cost of labeling while still maintaining performance. AL can be categorized into stream-based sampling and pool-based sampling. The former makes an independent
judgment on whether the incoming sample in the data stream needs to query the labels, while the latter chooses the best query samples based on the evaluation and ranking of the whole unlabeled pool~\cite{ren2021survey}. 
Currently, the uncertainty-based approach~\cite{ren2021survey}, which samples the most uncertain
samples to form a batch query set, is commonly used in AL applications. 
Information entropy~\cite{settles2009active} is a commonly used metric to measure the prediction uncertainty of a sample. To consider the data distribution of the samples, Exploration-P~\cite{yin2017deep} added a redundancy which is represented by the similarity between the current sample and the selected sample set to the entropy. However, unlike classification tasks which provide only one classification prediction for each sample, the TAL tasks generate both classification predictions and action-background predictions for each clip of a video segment. 
Therefore, directly applying the AL strategies may not lead to superior performance.

To ease the large-scale data dependence of current TAL methods, Heilbron et. al.~\cite{heilbron2018annotate} developed a learner to bootstrap the active selection function from the existing data. Estimated by a testing set, the selection function can be learned to pick samples that improve the attention module of a TAL model at the most. However, providing fully labeled annotations for the selected sample and the testing set also costs lots of human effort. 
Like other AL works, such an approach requires requesting labels from the oracle after every training iteration and it is impractical for the cloud oracle to provide labels to the local devices in real-time.
Therefore, in realistic on-device scenarios, it is desirable to develop a new sampling strategy that selects the most representative video segments only once at the beginning of the training process.

%% file: 3_motivation.tex
\section{Framework Overview}
\label{sec:motivation}

We first introduce a conventional W-TAL method as our baseline in Section \ref{sec:baseline}. Since this baseline method is designed for learning from pre-collected videos (organized as separate video files) and cannot be directly applied to learning from a video stream, we then present a stream learning workflow and discuss two challenges of learning from the video stream in Section~\ref{sec:case}, which will be addressed by our methods in Section~\ref{sec:segment} and Section~\ref{sec:sample}, respectively.
\vspace{-6pt}

\subsection{Revisiting Video Learning}
\label{sec:baseline}

In this paper, we adopt CoLA~\cite{zhang2021cola} as our baseline method. CoLA utilizes CL to distinguish between action and background clips during the training process. We also use its temporal action location (TAL) model as our DNN backbone.

\begin{figure}[h]
\vspace{-6pt}
\centering
\includegraphics[width=
0.9\linewidth]{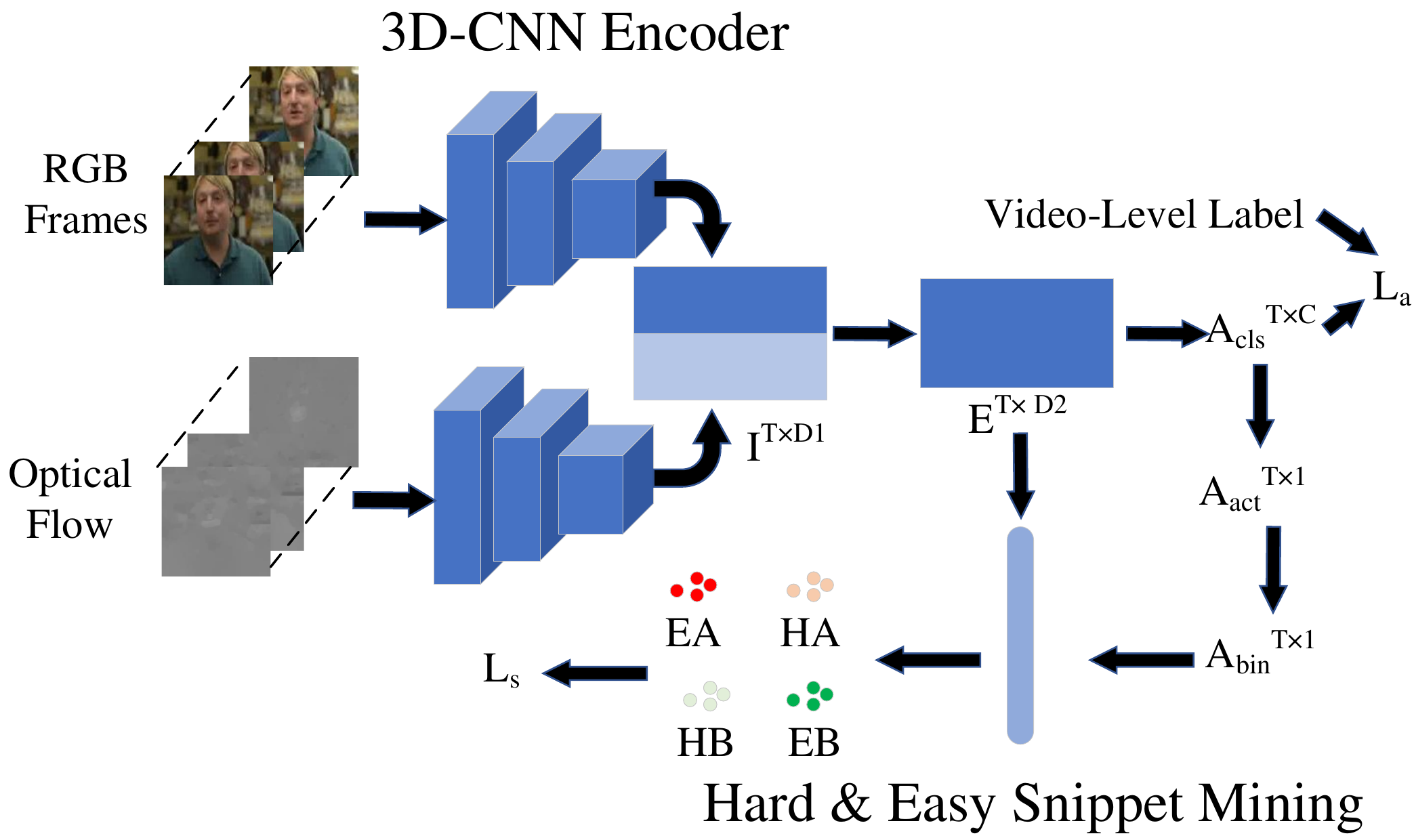}
\setlength{\abovecaptionskip}{-0.1cm}
\caption{Our baseline method CoLA~\cite{zhang2021cola}. It learns from pre-collected video files and cannot directly be applied to on-device learning from a video stream.}
\label{fig:cola}
\vspace{-6pt}
\end{figure}

The CoLA framework is shown in Fig.~\ref{fig:cola}.
In TAL tasks, a DNN model is learned from a set of untrimmed videos. In CoLA~\cite{zhang2021cola}, RGB frames and optical flow were extracted from each video with 25fps. The RGB frame stream and optical flow stream of a video were divided uniformly into non-overlapping clips where each clip included 16 frames. The encoder of a pre-trained 3D-convolutional neural network (3D-CNN), I3D, was used to extract RGB features and optical flow features. The RGB and optical flow features of each clip were concatenated, and a fixed number of $T$ clips of features were sampled as input features $I^{T\times D1}$ due to the variation of video length. The embedded features $E^{T\times D2}$ were extracted from $I^{T\times D1}$ by fully connected (FC) layers. Then, a classifier was applied to generate the Temporal Class
Activation Sequence (T-CAS) $A_{cls}^{T\times C}$, where $C$ was the number of action categories of the dataset. The T-CAS was summed along the channel dimension to obtain a class-agnostic aggregation actionness $A_{act}^{T\times 1}$. For localization, by comparing with a threshold, the action attention value for $A_{act}^{T\times 1}$ of each clip was converted into a binary value, and the binary value sequence $A_{bin}^{T\times 1}$ was used to distinguish actions and background. For classification, clips with top-k highest action attention values from $A_{act}^{T\times 1}$ were selected, and the mean of their classification results served as video-level predictions. During the training phase, the video-level labels were provided, and the action loss $L_a$ was calculated by the cross-entropy function. A Hard \& Easy Snipped Mining strategy was explored to select embedded features of EA, EB, HA, and HB, where EA and EB were from clips with top-k highest and lowest action attention values, respectively. HA and HB were from boundary-adjacent
clips that were hard to distinguish, but HA was located to be closer to EA, while HB was located to be closer to EB. Since HA was closer to EA, HA was more likely to be an action and should share similar embedded features with EA. Similarly, the embedded features of HB were also supposed to be similar to that of EB. Based on such observation, the SniCo loss $L_s$ was computed in~(\ref{eq:SniCo}). In~(\ref{eq:SniCo}), $E_{x\sim X^{HA},x^+\sim X^{EA},x^-\sim X^{EB}}l(x,x^+,x^-)$ represented the NCE loss between HA, EA, and EB. The NCE loss is a loss function used in CL, which can be computed in~(\ref{eq:NCE}). In~(\ref{eq:NCE}), $x$ and $x^+$ are extracted features that are considered to be similar. For example, in CL-based image classification tasks~\cite{chen2020simple,he2020momentum}, the inputs generated from the same image via different data augmentation should have similar features extracted by the CNN encoder. Such features are defined as positive pairs. Similarly, features extracted from different images should be distinguished. Such features are defined as negative pairs which are represented as $x$ and $x^-$. The temperature $\tau$ is a constant parameter that is selected as 0.07 in CoLA. CoLA aimed at distinguishing boundary-adjacent clips so that the embedded features of HA should be similar to the embedded features of EA, and the embedded features of HB should be similar to that of EB. Therefore, in~(\ref{eq:SniCo}), CoLA used the mean of embedded features (i.e. the mean of $E^{T\times D2}$ on the time dimension) of HA and the mean of embedded features of EA as positive pairs. The mean of embedded features of HA and all the EB embedded features were considered as negative pairs. Similarly, $E_{x\sim X^{HB},x^+\sim X^{EB},x^-\sim X^{EA}}l(x,x^+,x^-)$ represented the NCE loss between HB, EB, and EA, where the mean of HB embedded features and the mean of EB embedded features were considered as positive pairs, while the mean of HB embedded features and EA embedded features were considered as negative pairs.
\vspace{-6pt}

\begin{equation}
\vspace{-6pt}
\begin{aligned}
L_s=E_{x\sim X^{HA},x^+\sim X^{EA},x^-\sim X^{EB}}l(x,x^+,x^-)+\\ E_{x\sim X^{HB},x^+\sim X^{EB},x^-\sim X^{EA}}l(x,x^+,x^-)
\label{eq:SniCo}
\end{aligned}
\vspace{-6pt}
\end{equation}

\begin{equation}
\begin{split}
&l(x,x^+,x^-)=\\&-log(\frac{exp(x^T\cdot x^+/\tau)}{exp(x^T\cdot x^+/\tau)+\sum_{s=1}^{k}
exp(x^T\cdot x^-_s/\tau)})
\label{eq:NCE}
\end{split}
\vspace{-6pt}
\end{equation}

In the testing phase, the CoLA model first generated T-CAS and aggregated top-k classification scores to get the video-level prediction. Then, it selected candidate proposals (i.e. contiguous clips that may contain action) by picking up clips with action attention values and classification scores higher than a set of thresholds. Finally, non-maximum suppression
(NMS) was applied to remove duplicated proposals.
\vspace{-6pt}

\subsection{Overview of On-Device Learning from A Video Stream}
\label{sec:case}

Unlike CoLA~\cite{zhang2021cola} and other TAL works which are learned from a set of manually separated video files, we explore learning from a long and raw video stream directly captured by on-device cameras. 
The video stream contains all action and background instances without laborious manual splitting.

\begin{figure}[htb]
\vspace{-6pt}
\centering
\includegraphics[width=
0.80\linewidth]{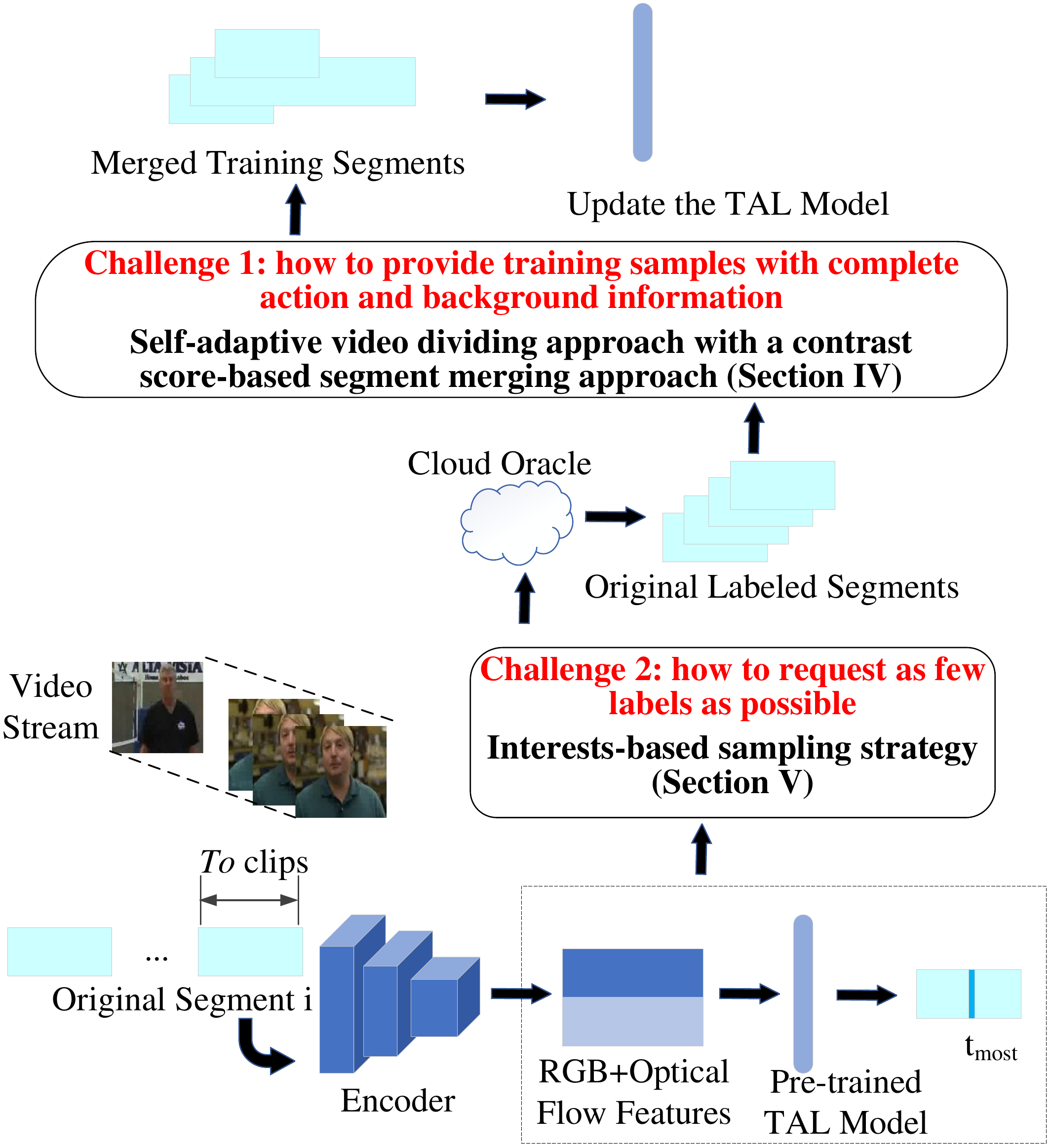}
\setlength{\abovecaptionskip}{-0.1cm}
\caption{Overview of our streaming learning workflow. We propose a self-adaptive video dividing approach with a contrast score-based segment merging approach to split the video stream as segments, each of which contains complete action and background frames for better CL. We then explore how to select the most representative video segments such that requesting their labels can benefit the learning most.}
\label{fig:workflow}
\vspace{-6pt}
\end{figure}


The proposed streaming learning workflow is shown in Fig.~\ref{fig:workflow} and it has three steps. In the \emph{first} step, the camera keeps collecting streaming data which will be stored directly on the local device. Considering our on-device camera collects hours of a video stream which can be converted to RGB frames and optical flows with 25 fps, and 16 frames compose a clip. The single stream is divided into non-overlapping segments uniformly with $To$ clips per segment. For the clips of each segment $i$, same as previous W-TAL works~\cite{paul2018w,zhang2021cola,su2021improving}, we use a frozen pre-trained I3D 3D-CNN encoder to extract features. We also adopt a pre-trained TAL model with the same backbone of CoLA to select the most representative clip $t_{most}$ for each segment, i.e. the clip with the top-1 highest action attention value. The selected clips will later be transmitted to the cloud oracle to request segment-level labels. It should be noted that since the camera keeps collecting streaming data for hours or days, it is inefficient to store all the raw video segments, extracted RGB frames, and optical flows on the device. 
We use the widely-used THUMOS’14 dataset to illustrate our method.
This dataset includes training videos with around 11 hours in total, which takes up over 15 GB of memory space to store the raw video and 70 GB of memory space to store the RGB frames and optical flow. 
Therefore, instead of storing raw inputs of segment $i$, only the extracted features and selected clips are stored in local memory. For the THUMOS’14 dataset, the extracted RGB and optical flow features take up only less than 2 GB of memory space in total, which is around $13\%$ of the original video and $3\%$ of the input RGB frames and optical flows.  
If $T=50$, the total length of the representative clips is only $2\%$ compared to the original video stream. Therefore, it is feasible for the local device to store all the extracted features and the most representative clips.

In the \emph{second} step, 
we select segments that should be labeled. Rather than sending the segment to the oracle, we only transmit the most representative clips selected by the pre-trained TAL model to the oracle to request segment-level labels of these segments to the cloud oracle. We assume the oracle with domain experts can provide the correct action class label for every selected clip, and these clip labels serve as weak, segment-level labels. We keep collecting the video during the daytime, and we upload the clips to the oracle every night.  The \emph{third} step begins after the oracle sends back all the labels. In this step, we use the weakly labeled segments to update the TAL model.

To effectively learn from the streaming video, there are two challenges to consider. The \textbf{first challenge} is how to pre-process the labeled segments and split them as effective training samples. Since the video stream is divided into segments with a pre-defined segment length during the video recording procedure, it is impossible to guarantee each segment contains a single complete instance. As mentioned in Section~\ref{sec:Introduction}, the divided segment should be neither too short nor too long. Since $To$ is pre-defined before the video stream features are collected, it is impossible to set an appropriate original segment length before the training start. To provide appropriate training samples that can benefit the training performance, we propose a self-adaptive dividing approach with a contrast score-based merging strategy to decide whether two continuous labeled segments which share the same action class labels need to merge or not so that the TAL model can learn from the new segments with complete action and background information. Such an approach will be discussed in detail in Section~\ref{sec:segment}.

The \textbf{second challenge} is how to select the video segments for labeling such that we can label as few labels as possible while maintaining the model accuracy. For example, for the THUMOS’14 dataset with around 11h video in total, if $To=50$, it will generate 1304 segments. Although we only need to send the most represent short clips to the cloud oracle, there are still 1304 clips that the oracle needs to deal with. In fact, some segments play little role in improving learning performance. Therefore, it is necessary to sample the most important segments based on the labeling budget of the oracle. AL learning is a method to deal with such problems. However, in AL, the learner requests labels from the oracle after every training iteration. It requires dedicated experts to label the segments from time to time as long as the system is operating the learning process. However, in on-device learning applications, it is impossible for the cloud oracle to generate labels for the local device in real-time. Therefore, a new data sampling strategy is required which can select the most useful segments to be labeled before the whole training process begins. We will explore different sampling strategies on TAL tasks and propose an interests-based sampling strategy that selects the segments that contain more interesting areas. The discussion will be illustrated in detail in Section~\ref{sec:sample}.

%% file: 4_algorithm.tex
\section{Self-adaptive Video Dividing Approach}
\label{sec:segment}

In this section, we address the first challenge discussed in Section~\ref{sec:case}. As explained in Section~\ref{sec:Introduction}, dividing the segment too short or too long will degrade the performance of the learned model. We propose techniques to split the video stream into an appropriate length of segments that can benefit the training performance most.  In this section, we assume all the segment labels are available, and the second challenge of labeling will be addressed in Section \ref{sec:sample}.

\begin{figure}[h]
\vspace{-6pt}
\centering
\includegraphics[width=
0.8\linewidth]{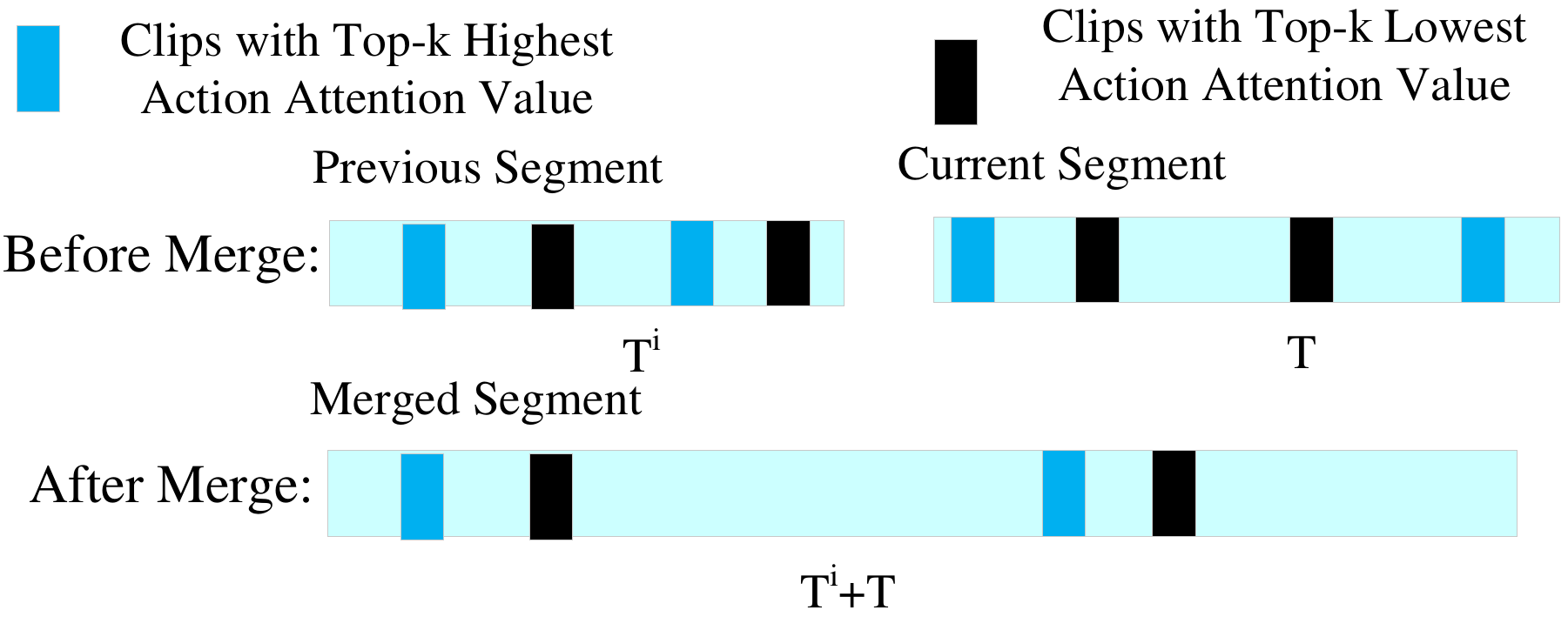}
\setlength{\abovecaptionskip}{-0.2cm}
\caption{Contrast score-based segment merging approach.}
\label{fig:merge}
\vspace{-6pt}
\end{figure}

In our proposed stream learning workflow, the single and long video stream is divided into non-overlapping segments uniformly with $To$ clips per segment. Assume the cloud oracle can provide labels for all the representative clips of the segments. The label for one representative clip is considered as the weak label of the whole segment. Therefore, during the training process, we can merge adjacent segments sharing the same weak label in a self-adaptive manner. Inspired by~\cite{wu2021enabling} which utilized the contrast score to select representative training samples from streaming images during contrastive learning, we propose a contrast score-based segment merging (CS-M) approach. However, unlike~\cite{wu2021enabling} which exploited the contrast score to measure whether the current image has been well learned by the model, our approach uses the contrast score to judge whether two contiguous video segments need to be merged or not so that they can provide complete action and background information for the CL-based TAL model. Therefore, the selection of contrast pairs should be based on the TAL framework. The CS-M is illustrated in Fig.~\ref{fig:merge}. 

As shown in Fig.~\ref{fig:merge}, the video stream is divided into segments where each segment contains $To$ contiguous non-overlapping clips  together. In this section, we assume all the representative clips of the segments can be labeled in the \emph{second} step of the stream learning workflow, i.e. all the segments can acquire segment-level labels. Starting from the first segment, for two adjacent segments sharing the same segment-level label, we select clips  $\{ca_j\}_{j=1,...,k}$ with the top-k highest action attention value (i.e. the clips that are predicted as easy actions by the current TAL model) and $\{cb_j\}_{j=1,...,k}$ with the top-k lowest action attention value (i.e. the clips that are predicted as easy background by the current TAL model). Following previous weakly-labeled TAL approaches~\cite{paul2018w,zhang2021cola}, $k=max\{1,\lfloor\frac{T^i}{s}\rfloor\}$, where $T^i$ is the length of the $ith$ segment and $s$ is a hyperparameter. $T^i$ may not equal to $To$ for the previous segment since it may be merged with the one before it, while the current segment which has not been merged has the original length $To$. Then, we decide whether to merge the previous segment and the current segment together by comparing the mean of the contrast score of the two segments before merging and the score of the segment after merging. 
\begin{equation}
\begin{split}
& contrast \ score \\  =  & sim\{mean\{eca_j\}_{j=1,...,k},mean\{ecb_j\}_{j=1,...,k}\}
\label{eq:cs}
\end{split}
\vspace{-6pt}
\end{equation}

 The contrast score is shown in~(\ref{eq:cs}), where $eca_j$ and $ecb_j$ are embedded features of $ca_j$ and $cb_j$ extracted from the current TAL model, respectively. The $sim\{a,b\}$ operation calculates the similarity of the two vectors $a$ and $b$ by the dot product operation of the normalized vectors. In CoLA, the EA and EB are set as two anchors. By enabling HA to be closer to EA and HB to be closer to EB, the actions and backgrounds can be distinguished explicitly. Therefore, we use the contrast score to measure the similarity of EA and EB. If the score is high,  i.e. EA and EB are more similar, it means that the segment does not have explicit EA and EB anchors, thus the segment may not contain complete action and background information to improve the TAL model. If the contrast score is low, it means that the segment can provide explicit EA and EB anchors for the model TAL to learn. Therefore, if the mean of the contrastive scores of the previous segment and current segment is higher than that of the merged segment, we merge the two segments. If the value of the merged segment is higher, it means the segments before merging already include explicit EA and EB, thus they have complete action and background information and do not need to be merged. After we decide whether the previous segment needs to be merged with the current one, we will move to the next segment. This merging operation happens before each training epoch. When the model improves during the training process, it can make more precise predictions of EA and EB, so our merging approach is self-adaptive during the learning process.

%% file: 5_strategy.tex
\section{Studies of Different Sampling Strategies}
\label{sec:sample}

In Section~\ref{sec:segment}, we assume the cloud oracle can provide labels for all segments. However, assuming we set $To$ as 50, there are still 1304 segments that need to be labeled. Therefore, it is necessary to sample the most important segments based on the labeling budget of the oracle. Previous AL works~\cite{yin2017deep,heilbron2018annotate} requested labels during every training iteration. However, such an AL process requires dedicated experts from the cloud center to label the segments from time to time as long as the system is operating the training process, so it is infeasible for the cloud oracle to provide labels in real-time. Therefore, we need to find an efficient sampling strategy to select the most useful segments to be labeled before the whole training process begins.

\begin{figure}[h]
\vspace{-6pt}
\centering
\includegraphics[width=
0.8\linewidth]{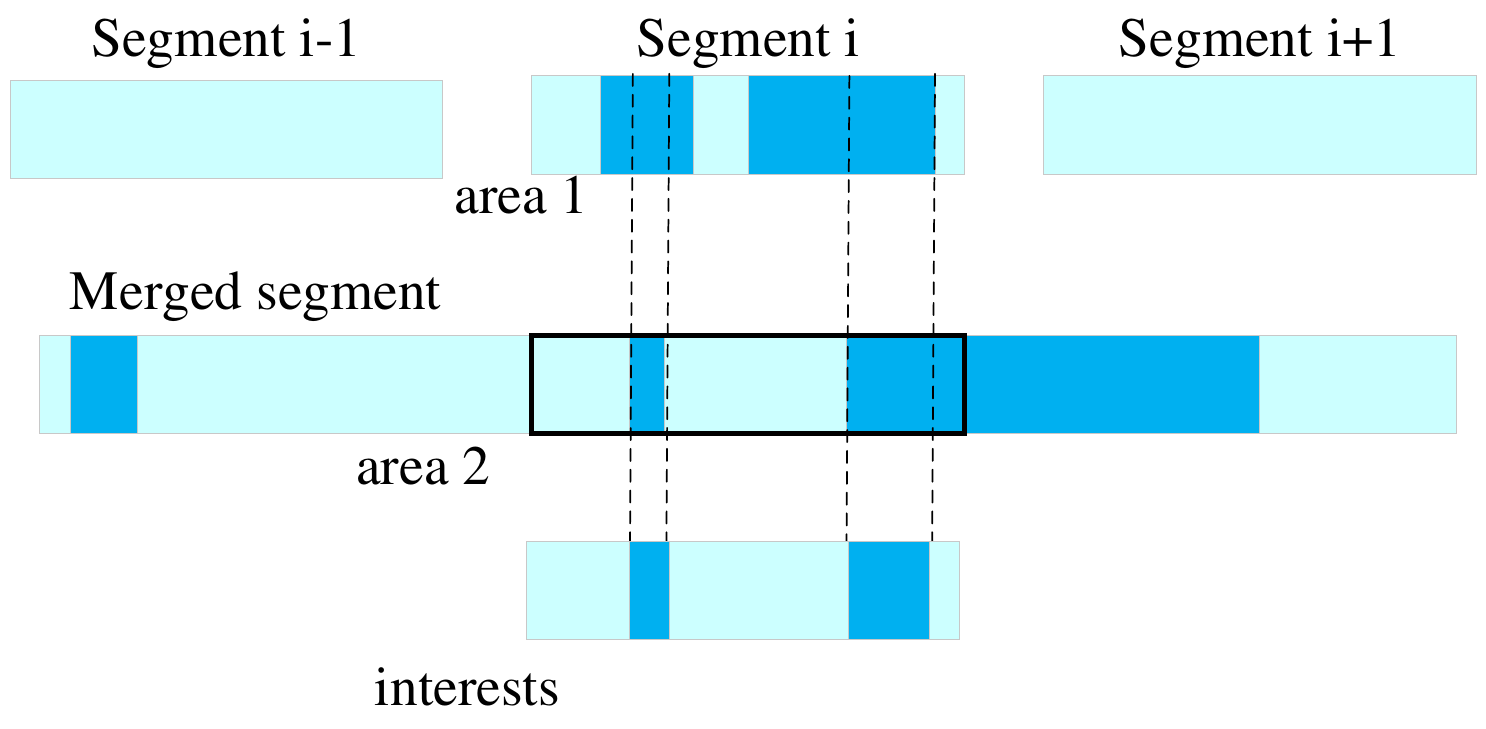}
\setlength{\abovecaptionskip}{-0.2cm}
\caption{Interests-based sampling. Areas of interests are sampled such that labeling them will benefit the model learning most.}
\label{fig:interests}
\vspace{-6pt}
\end{figure}

In on-device scenarios, the video stream always contains very long background periods. Some divided segments may contain only a few action clips. Therefore, we propose an interests-based sampling (IS) strategy that selects segments that have more interesting areas (i.e. the areas that are more likely to contain action). The sampling strategy is shown in Fig.~\ref{fig:interests}.

For segment $i$, we use the pre-trained model to find action proposals using the same detection process of the CoLA baseline. The only difference is that when applying NMS, we set the intersection over the union (IoU) threshold to be zero so that the selected proposals are non-overlapping. The selected proposals are represented as $area1$. It is noted that the adjacent segments are more likely to have the same action instances. Therefore, we generate another prediction by merging segment $i$ with segment $i-1$ and segment $i+1$. We utilize the same way to predict non-overlapping proposals of the merged segment and the predicted non-overlapping proposals in segment $i$ are represented as $area2$. The intersect proposals of $area1$ and $area2$ are represented as $interests$ which means that these areas are more likely to have actions since they are predicted as action proposals from two testing procedures. The selected metric is formulated in (\ref{eq:interests}). Then, we select the segments with the top-$n\times c$ score, where $c$ is the number of action categories, and $n\times c$ is the number of segments we select to request labels. 

 \begin{equation}
\begin{aligned}
score_i=\frac{\text{length\ of\ interests}}{\text{length\ of\ segment\ i}}
\label{eq:interests}
\end{aligned}
\end{equation}

It should be noted that the sampled segments may not be consecutive to each other. Therefore, in this step, we also keep the indexes of the sampled segments in the original stream. Then, in the merging process in Section~\ref{sec:segment}, we only consider whether to merge two consecutive segments if they share the same segment-level labels.

%% file: 6_exp.tex
\section{Experiments}
\label{sec:experiments}

\subsection{Experimental Setup}
\label{sec:setup}
\textbf{Dataset and Baseline} We use the THUMOS’14 dataset which includes untrimmed videos with 20 categories~\cite{idrees2017thumos}. We use the W-TAL work CoLA~\cite{zhang2021cola} as our baseline and use the same TAL model as the backbone which includes 2 1D convolution layers as the classifier. Same with the baseline, we use the 200 videos in the validation set for training and the 210 videos in the testing set for evaluation. Unlike the W-TAL baseline which trains the model from separate untrimmed videos, we combine the 200 videos together with the following orders to form different single, long video streams. (1) The videos are combined randomly with two different random seeds. (2) The videos are combined randomly and at least two consecutive videos in the input stream are from the same class. (3) The videos are combined in the original order of the dataset (i.e. the videos from the same class are consecutive in the input stream). The different combination orders represent different temporal correlations of the input stream. The more videos from the same action category are consecutive means the temporal correlation of the stream is stronger~\cite{wu2021enabling}. We use the combined stream to reflect the behavior of a video stream collected in a real-life environment because they share the following features. Firstly, the combined stream lasts for several hours, and the actions of nearly all classes are included in this stream. Secondly, background clips are between two adjacent action instances. Thirdly, two adjacent action instances are likely to be from both the same action categories and the different categories based on the temporal correlations of the input stream.

\textbf{Pre-training and Feature Extraction} Same with our baseline, we use the I3D network pre-trained on the Kinetics dataset as our encoder~\cite{encoder}. RGB frames and optical flows are extracted from the raw video, and the video stream is divided into 16-frame non-overlapping clips. For each clip, the RGB features and optical flow features which both have 1024 channels are concatenated together. To select the most representative clips of each segment and sample the most representative segments to request labels, we pre-train the classifier of the TAL model on the UCF-101 dataset~\cite{soomro2012ucf101}. It has 9537 trimmed videos from 101 categories and 1966 of them share the same action categories as that in our target THUMOS'14 dataset. Therefore, we use the 1966 trimmed videos to pre-train the classifier of the TAL model. To make a fair comparison with our baseline, we use the same extracted concatenated features available from the codes provided by CoLA~\cite{CoLAcode} for the target THUMOS'14 dataset.

\textbf{Default Training Setting} To make a fair comparison with the W-TAL baseline, we use the same hyper-parameters according to the codes of CoLA~\cite{CoLAcode}, such that the number of EA or EB is $T/5$, and the number of HA or HB is $T/20$. In CoLA, $T$ is set to 750 for the individual untrimmed videos. In our work, we set $T$ to the median length of all labeled merged segments. The batch size is 16, and the learning rate is 0.0001, which are same as the baseline. We train the TAL model for 500 epochs. 

\textbf{Evaluation Protocols} To make a fair comparison, we use the same testing code and thresholds in the baseline~\cite{CoLAcode} to detect action proposals on the testing set. Then, we follow the standard evaluation
protocol by reporting mean Average Precision (mAP) values when the IoU thresholds are set to $[0.1:0.7:0.1]$.
\vspace{-6pt}

\subsection{Experimental Results of Contrast Score-based Segment Merging Approach}
\label{sec:merge_result}
In this section, we validate the efficiency of our contrast score-based segment merging approach in our self-adaptive video dividing approach. When the video stream is coming, we divide them into original segments evenly with the length of $To$. We divide the total number of clips of the whole stream by $To$ to calculate the number of original segments (NoOS). In our experiments, $To$ is set to 25, 50, 100, and 150, respectively. We select the most representative clip for each segment to request a segment-level weak label. In our experiment, we consider the label of the action segment that is closest to the selected clip as the label of the clip. The original videos of the dataset are combined into a long video stream in four different orders: the videos are combined randomly with seed=0, the videos are combined randomly with seed=10, the videos are combined randomly and at least two consecutive videos in the input stream are from the same class, and all the videos from the same class are consecutive in the input stream. In each order, we train the TAL model both transferred from the model pre-trained on the trimmed UCF-101 dataset and trained from scratch. We calculate the average testing mAP when the IoU threshold ranges from 0.1 to 0.7 and then calculate the mean of the average testing mAP scores under the four different streaming video orders.

\textbf{Baselines}
We compare CS-M with several video merging strategies, including \emph{without merging} (WM), \emph{random merging} (RM), and \emph{merging all} (MA). The WM strategy means that we use the original segments that are uniformly divided from the whole stream without any pre-processing as the training samples. The RM strategy means that for the original segments that share the same label, we randomly divide these segments into randomly set groups where the segments in each group are continuous. Then, we merge the segments that are from the same group together. The MA strategy means the adjacent segments are merged as long as they share the same segment-level label. For the CS-M strategy, we test merging for one iteration (1 iter), two iterations (2 iters), and three iterations (3 iters) in each epoch, respectively. We also try to split the segment after a contrast score-based merging iteration (CS-MS). We first determine the split point of a segment by finding two consecutive clips having the lowest cosine similarity. Then, we use the contrast score to decide whether the segment needs to be split or not. We also compare our results with that of CoLA~\cite{zhang2021cola}. It represents dividing the whole stream into separate video segments with exhaustive effort and training the model with the well-divided videos under the W-TAL setting.

\textbf{Results}
The comparison of different merging strategies when training the TAL model from the model pre-trained on the trimmed UCF-101 dataset is shown in Table~\ref{tab:avgmerge_mAP_pre}, while the result of the merging strategies when training the TAL model from scratch is shown in Table~\ref{tab:avgmerge_mAP_scratch}. As illustrated in Table~\ref{tab:avgmerge_mAP_pre} and Table~\ref{tab:avgmerge_mAP_scratch}, conducting three iterations of contrast score-based merging, i.e. CS-M (3 iters), in a training epoch achieves superior testing performance in general when $To$ is 25, while CS-M (2 iters) works well when $To=50$ when training from the pre-trained model. In the rest situations, CS-M (1 iter) outperforms other strategies. When $To=50$, the CS-M (1 iter) strategy achieves the highest testing performance with 37.55\% mAP when trained from the pre-trained model and 37.51\% mAP when trained from scratch.

\begin{table}[!t]
\setlength\tabcolsep{3pt}
{\caption{Comparison of the mean of average Testing mAP (\%) scores of the merging strategies under four different streaming video orders when training from the pre-trained model \label{tab:avgmerge_mAP_pre}
}}
\centering
\begin{tabular}{|c|c|c|c|c|c|c|c|c|}
\hline
 $To$ & NoOS 
 &WM &RM
&MA &\makecell[c]{CS-M\\(1 iter)} &\makecell[c]{CS-M\\(2 iters)} &\makecell[c]{CS-M\\(3 iters)} &CS-MS

 \\
\hline
25 & 2608
&27.43  &28.99 &36.08 &35.19 &35.86 &\textbf{36.33}  &33.78

\\ \hline
50 & 1304 &	31.42  & 31.90	 & 35.96  &37.55 &\textbf{37.62} &36.84 &35.63

\\

\hline
100	& 652 &	34.89  &34.26  &34.97  &\textbf{36.73} &35.40 &35.17 &35.69
\\
\hline

150	 & 435	&35.99  &	36.03  & 34.79  &\textbf{36.32} &35.48 &35.83 &35.04\\
\hline

\end{tabular}
\vspace{-6pt}
\end{table}

\begin{table}[!t]
\setlength\tabcolsep{3pt}
{\caption{Comparison of the mean of average Testing mAP (\%) scores of the merging strategies under four different streaming video orders when training from scratch \label{tab:avgmerge_mAP_scratch}
}
}
\centering
\begin{tabular}{|c|c|c|c|c|c|c|c|c|}
\hline
 $To$ & NoOS 
 &WM &RM
&MA &\makecell[c]{CS-M\\(1 iter)} &\makecell[c]{CS-M\\(2 iters)} &\makecell[c]{CS-M\\(3 iters)} &CS-MS

 \\
\hline
25 & 2608
& 27.87 & 28.90 &35.80 & 35.28 &36.24 &\textbf{37.25} &34.55

\\ \hline
50 & 1304 &	31.90 & 32.56 &36.03 &\textbf{37.51} &36.24 &36.74 &36.16
\\

\hline
100	& 652 &	34.82 &	35.26 &	34.75 &	\textbf{36.71} & 35.97 &35.87 &35.23
\\
\hline

150	 & 435	& 35.74 &	35.94 & 34.73 &	\textbf{36.25} &35.56 &34.96 &35.85
\\
\hline

\end{tabular}
\vspace{-6pt}
\end{table}

To better understand the performance of the four merging strategies, we discuss the learning results under the above-mentioned situations in detail. The results are as follows.

\textbf{Situation (1):} The videos are combined randomly.

\begin{figure}[h]
\vspace{-6pt}
\centering
\includegraphics[width=
0.88\linewidth]{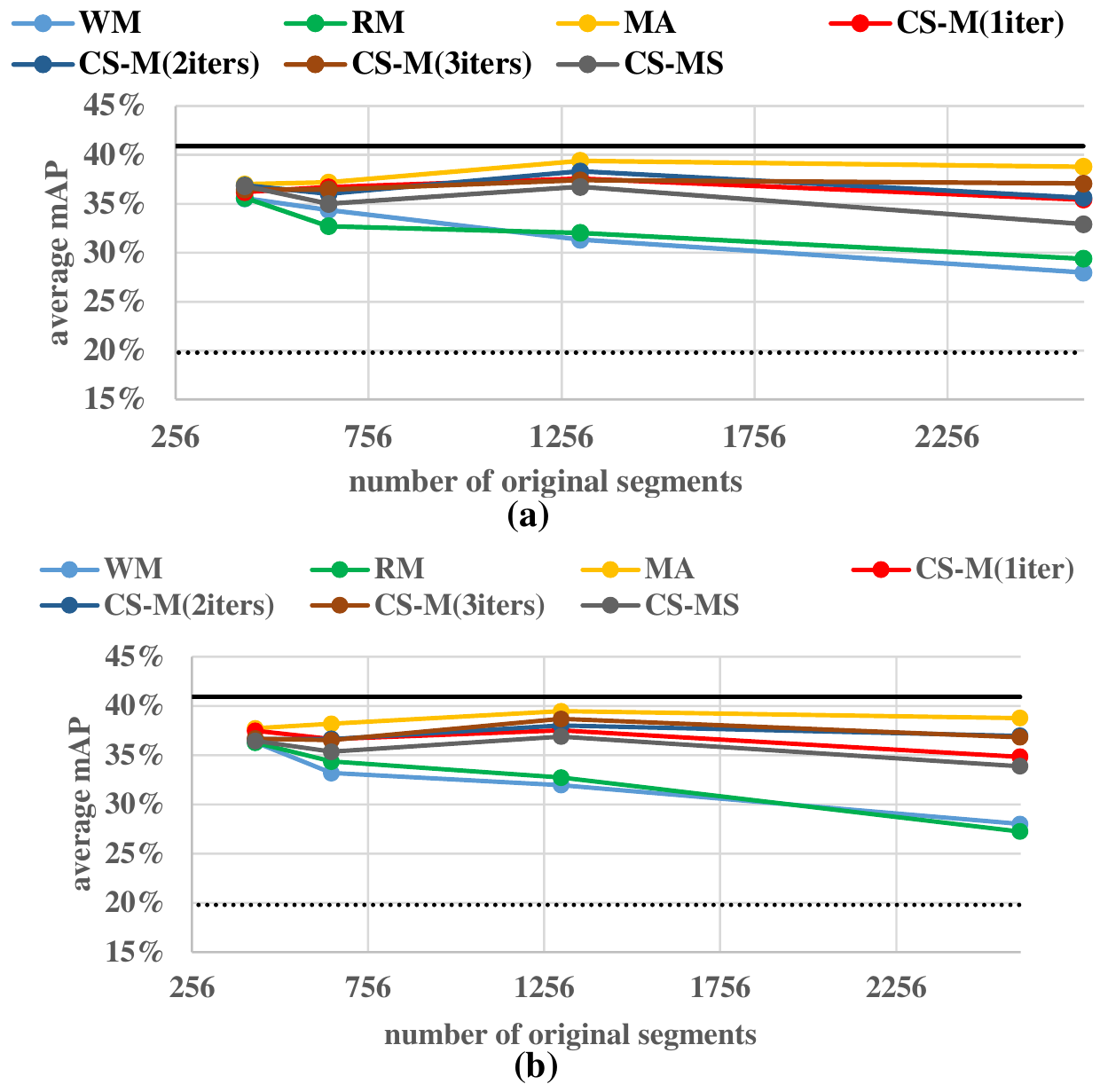}
\setlength{\abovecaptionskip}{-0.2cm}
{\caption{Comparison of different merging strategies in Situation (1) (seed=0). (a) Training the TAL model from the model pre-trained on the trimmed dataset. (b) Training the TAL model from scratch.}
\label{fig:mergeseed0}
}
\vspace{-6pt}
\end{figure}

\begin{figure}[h]
\centering
\includegraphics[width=
0.88\linewidth]{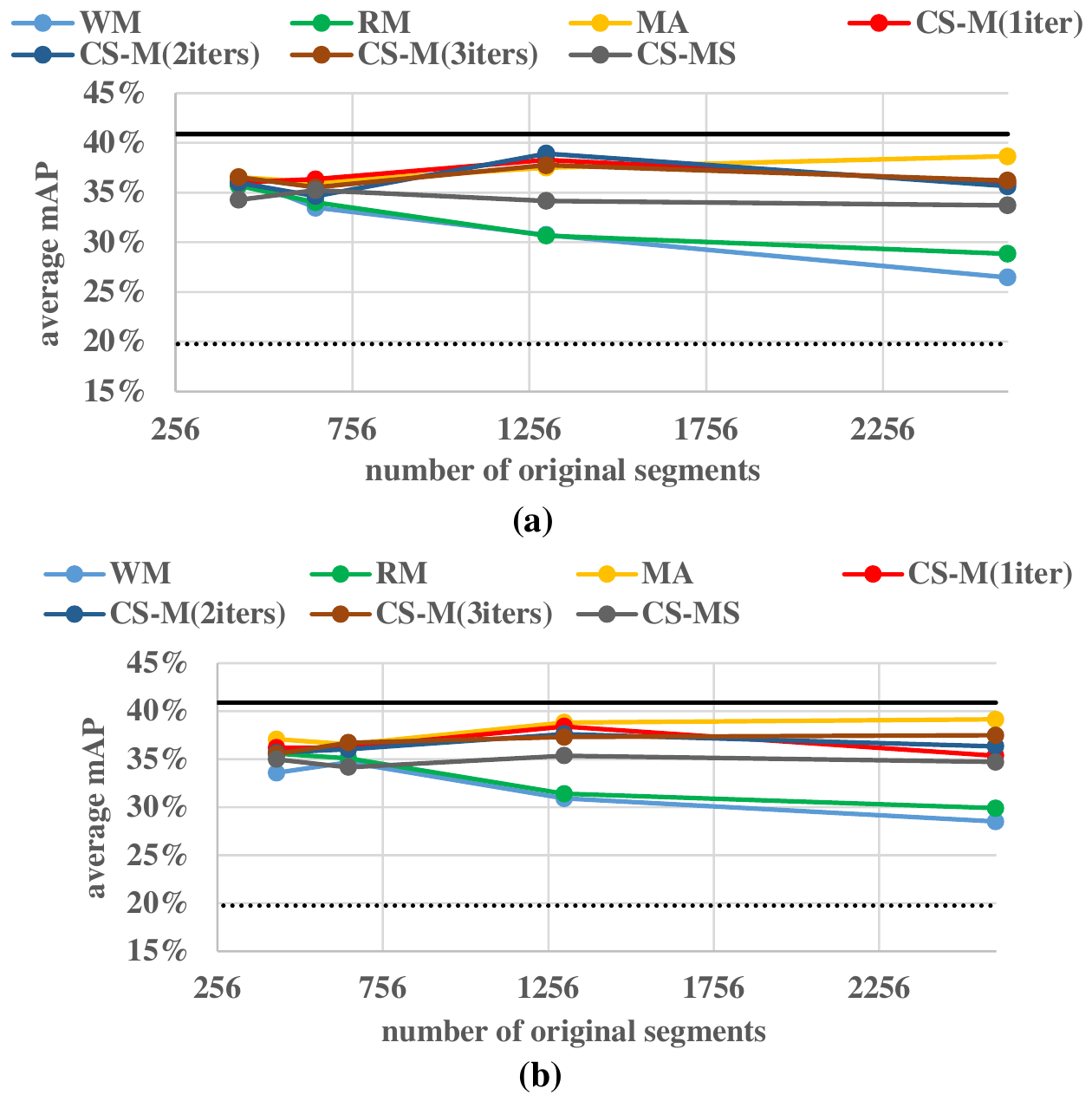}
\setlength{\abovecaptionskip}{-0.2cm}
{\caption{Comparison of different merging strategies in Situation (1) (seed=10). (a) Training the TAL model from the model pre-trained on the trimmed dataset. (b) Training the TAL model from scratch.}
\label{fig:mergeseed10}
}
\vspace{-6pt}
\end{figure}

The experimental results of training the TAL model from the video stream when the videos are combined randomly (with the random seed as 0 and 10) to form the stream are shown in Fig.~\ref{fig:mergeseed0} and Fig.~\ref{fig:mergeseed10} respectively. We compare the average testing mAP when the IoU threshold ranges from 0.1 to 0.7 of the proposed contrast score-based merging strategy with the results from the WM, RM, and MA strategies. The upper bound performance (the black solid line) is 40.9\% reported from CoLA~\cite{zhang2021cola} when the model is trained from separately untrimmed videos under the W-TAL setting. The lower bound (the black dash line) is the average testing mAP of the model pre-trained on the UCF-101 trimmed videos, which is 19.78\%. Fig.~\ref{fig:mergeseed0} (a) and Fig.~\ref{fig:mergeseed10} (a) show the results when the TAL model is transferred from the pre-trained model. The proposed CS-M strategy with different merging iterations outperforms the WM and RM strategies. It shows that the originally divided segments cannot convey complete action and background information for the TAL model to learn, so it is necessary to merge adjacent segments. Compared with RM, the proposed CS-M approach can judge whether the current segment already contains complete information and whether it needs further merging or not. The outcomes of CS-M (1 iter), CS-M (2 iters), and CS-M (3 iters) are similar but CS-M (3 iters) achieves a bit higher mAP when $To=25$. It shows that improving the merging iterations can improve the learning performance when the original segment is short, but the overall difference is small. For CS-M (1 iter), the best average testing mAPs are 37.58\% and 38.28\% when $To=50$. The performance of CS-MS lies between CS-M and WM, which means that such a splitting strategy cannot improve the learning performance.
 
The MA strategy is a bit better than the CS-M strategy with the best average testing mAP as 39.39\% when $To=50$ and 38.65\% when $To=25$. It is because when the original videos are randomly shuffled and combined into the video stream, it is more likely that the consecutive segments that share the same label are all generated from one video file of the original dataset. Therefore, MA is pretty close to the original W-TAL situation. However, the temporal correlation of the input stream is unknown during the training process, so we also consider the situations when the original videos are combined with the stream if consecutive videos are from the same class.

Fig.~\ref{fig:mergeseed0} (b) and Fig.~\ref{fig:mergeseed10} (b) show the results when the TAL model is trained from scratch. The performance is similar to that of training from the pre-trained model. The CS-M (1 iter) strategy with the best average testing mAP reaches 37.53\% and 38.41\% when $To=50$.

\textbf{Situation (2):} The videos are combined randomly and at least two consecutive
videos in the input stream are from the same class. 

\begin{figure}[h]
\centering
\includegraphics[width=
0.88\linewidth]{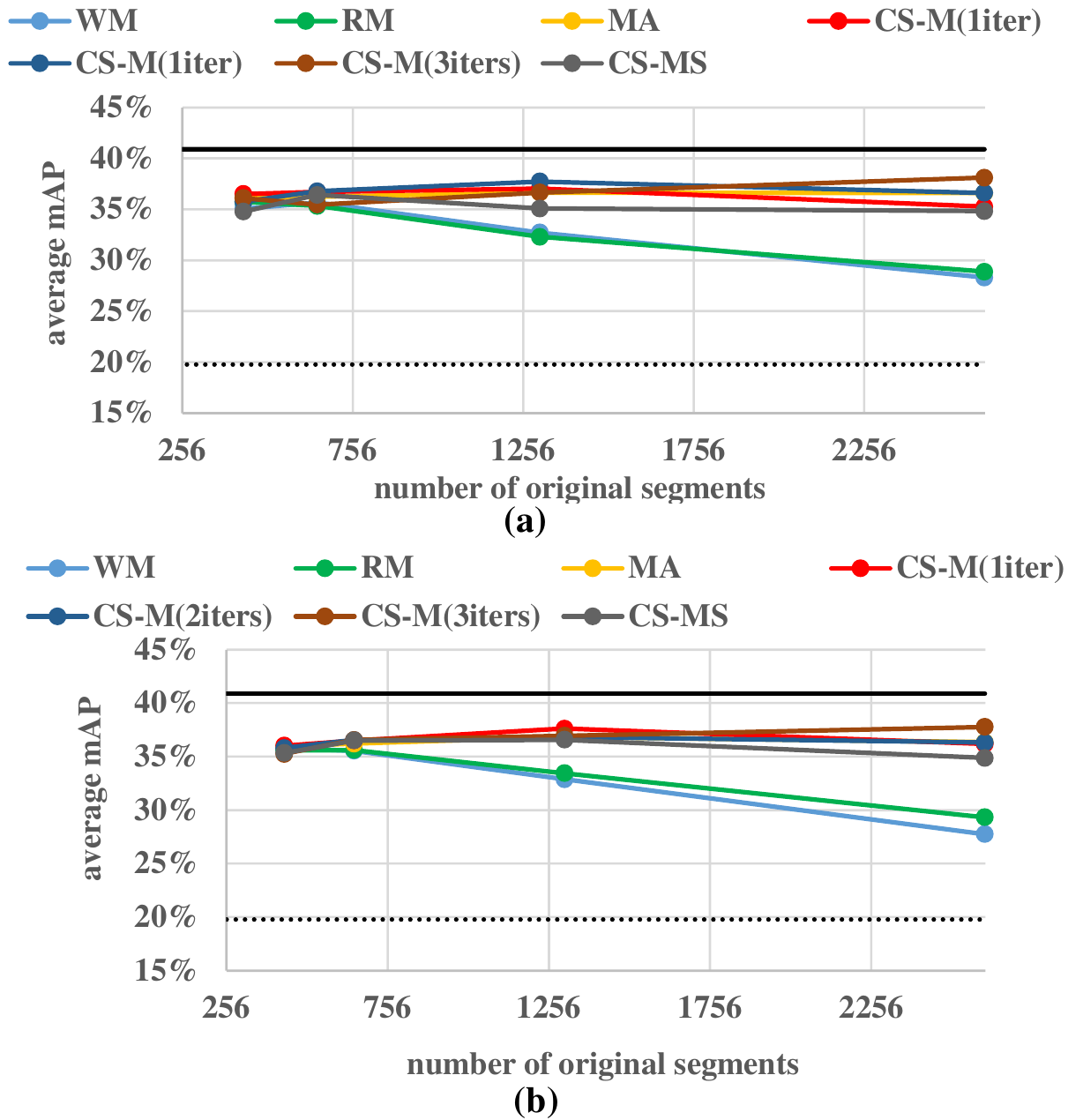}
\setlength{\abovecaptionskip}{-0.2cm}
{\caption{Comparison of different merging strategies in Situation (2) (seed=0). (a) Training the TAL model from the model pre-trained on the trimmed dataset. (b) Training the TAL model from scratch.}
\label{fig:mergecls}
}
\vspace{-6pt}
\end{figure}

This situation happens when continuous action instances are correlated to each other. As illustrated in Fig.~\ref{fig:mergecls}, when the original videos are randomly shuffled to form an input stream and at least two consecutive videos are from the same class, CS-M outperforms other merging strategies. The outcomes of CS-M (1 iter), CS-M (2 iters), and CS-M (3 iters) are close to each other in general. The best average testing mAP of CS-M (1 iter) reaches 37.04\% ($To=50$) and 37.63\% ($To=50$) when the TAL model is transferred from the pre-trained model and is directly trained from scratch, respectively. 

Compared to Situation (1), the performance of MA decreases a bit, because the merged segments are too long that contain action and background information from multiple instances which are hard to learn. Besides, if the length of merged segments elongates, there will be fewer learning samples in the same hours of the video stream. 

\textbf{Situation (3):} All videos from the same class are consecutive in the input stream. 

\begin{figure}[h]
\vspace{-6pt}
\centering
\includegraphics[width=
0.88\linewidth]{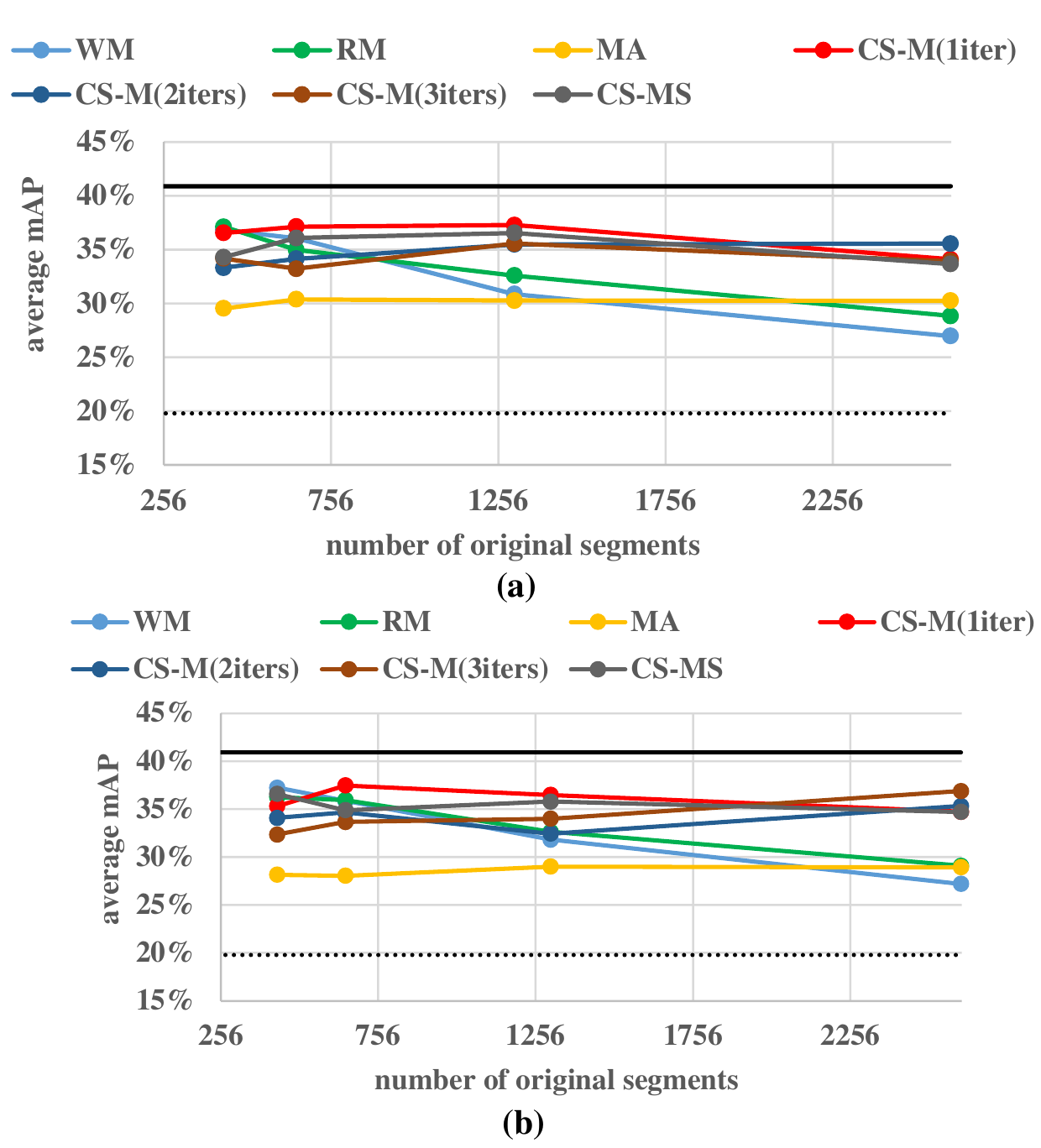}
\setlength{\abovecaptionskip}{-0.2cm}
{\caption{Comparison of different merging strategies in Situation (3). (a) Training the TAL model from the model pre-trained on the trimmed dataset. (b) Training the TAL model from scratch.}
\label{fig:mergeseq}
}
\vspace{-6pt}
\end{figure}

Finally, we combine the videos into a stream using the original order of the dataset so that the videos from the same class are consecutive in the input stream. Such a situation with a strong temporal correlation is also very common in on-device learning scenarios when the local devices keep receiving a series of action instances from one category in a fixed scene and then enter a new scene receiving action instances from another category. 

As illustrated in Fig.~\ref{fig:mergeseq}, the outcomes of CS-M (2 iters) and CS-M (3 iters) lie between that of CS-M (1 iter) and MA. In general, merging with fewer iterations per epoch works well when $To$ is larger while merging with more iterations per epoch works well when $To$ is small. For CS-M (1 iter), the best average testing mAP of the CS-M strategy reaches 37.28\% ($To=50$) and 37.47\% ($To=100$) when the TAL model is transferred from the pre-trained model and is directly trained from scratch, respectively.  

The average testing mAP of MA is around 30\% when learning from the pre-trained model, and below 30\% when learning from scratch. The average testing mAP decreases by more than 6\% compared to that in Situations (1) and (2). Therefore, the TAL model cannot learn from the merged segments under the current situation. It is because the merged segments are too long for the model to improve, while the number of samples is too small. For example, when $To=50$, the original segments are merged into only 23 samples, while the average length of the segments is 2834. Therefore, the MA strategy can only work when the temporal correlation of action instances is not strong.

\begin{table*}[!t]
\caption{Testing mAP of the contrast score-based merging strategy when $To=50$\label{tab:csmerge_mAP}}
\centering
\begin{tabular}{|c|c|c|c|c|c|c|c|c|}
\hline
  \multirow{2}{*}{Situation}
 &\multicolumn{8}{c|}{mAP@IoU (\%)}\\ \cline{2-9}
 &0.1&0.2&0.3&0.4&0.5&0.6&0.7&AVG\\
\hline
 (1) seed=0 & 61.86 (62.35) &  55.08 (55.33)&  47.07 (46.57)&  38.86 (38.64) &  29.81 (29.91)&  19.60 (19.71)&  10.74 (10.21)&  37.58 (37.53)\\ \hline

(1) seed=10 & 62.23 (62.41)&  56.03 (55.64)&  47.73 (47.78)&  39.94 (40.17)&  30.82 (30.88)&  20.75 (21.04)&  10.48 (10.97)&  38.28 (38.41)\\ \hline

(2) & 60.21 (61.30)&  53.99 (55.25)&  46.28 (47.11)&  38.68 (38.73)&  29.90 (30.21)&  19.71 (19.70)&  10.54 (11.11)&  37.04 (37.63)\\ \hline

(3) & 60.89 (60.21)&  54.89 (53.93)&  47.12 (46.20)&  38.82 (37.64)&  29.24 (28.46)&  19.99 (19.19)&  10.02 (9.76)&  37.28 (36.48)\\ \hline

W-TAL~\cite{zhang2021cola} & 66.2 &59.5 &51.5 &41.9 &32.2 &22.0 &13.1 &40.9\\
\hline
\end{tabular}
\vspace{-6pt}
\end{table*}

However, the temporal correlation is unknown during the training phase and $To$ is pre-defined before the video stream is collected, so it is desirable that the proposed approach can provide promising results under different situations with a fixed $To$. Table~\ref{tab:csmerge_mAP} shows the testing mAP of the proposed CS-M (1 iter) strategy learning from the input streams under the above-mentioned situations with $To=50$. 
The values inside the parentheses are the testing results of the TAL model trained from scratch, while the values outside the parentheses are the results of the model trained from the pre-trained model. The baseline, CoLA~\cite{zhang2021cola} was trained under the W-TAL setting, where the TAL model is learned from 200 separate untrimmed videos with video-level labels. As shown in Table~\ref{tab:csmerge_mAP}, the performance of our approach under the streaming learning setting is comparable with the W-TAL baseline. The average testing mAP decrease ranges from 4.42 to 2.49.

\vspace{-6pt}
\subsection{Experimental Results of Different Merging and Sampling Strategies}
\label{sec:sample_result}

In Section~\ref{sec:merge_result}, we assume the cloud oracle can provide labels for all original segments. However, for example, when $To=50$, there are 1304 clips requiring labels. If the oracle cannot provide all labels for the segments, it is desirable to sample the most representative segments for labeling. Unlike AL which requests labels after one training iteration or epoch, it is infeasible for the oracle to provide labels to the device in real-time. Therefore, in our experiments, the original segments are sampled after the input stream is recorded and all labels of the clips of selected segments are given at the beginning of the training process.

The labeling budget ranges from 1c to 30c, while c is the number of action categories. We consider the three situations the same as that in Section~\ref{sec:merge_result}.

\begin{equation}
\begin{aligned}
entropy_i=mean\{-(A_{ij}\times log(A_{ij})+\\(1-A_{ij})\times log(1-A_{ij})\}_{j=1,...,To}
\label{eq:us}
\end{aligned}
\vspace{-6pt}
\end{equation}

\begin{figure}[h]
\vspace{-6pt}
\centering
\includegraphics[width=
\linewidth]{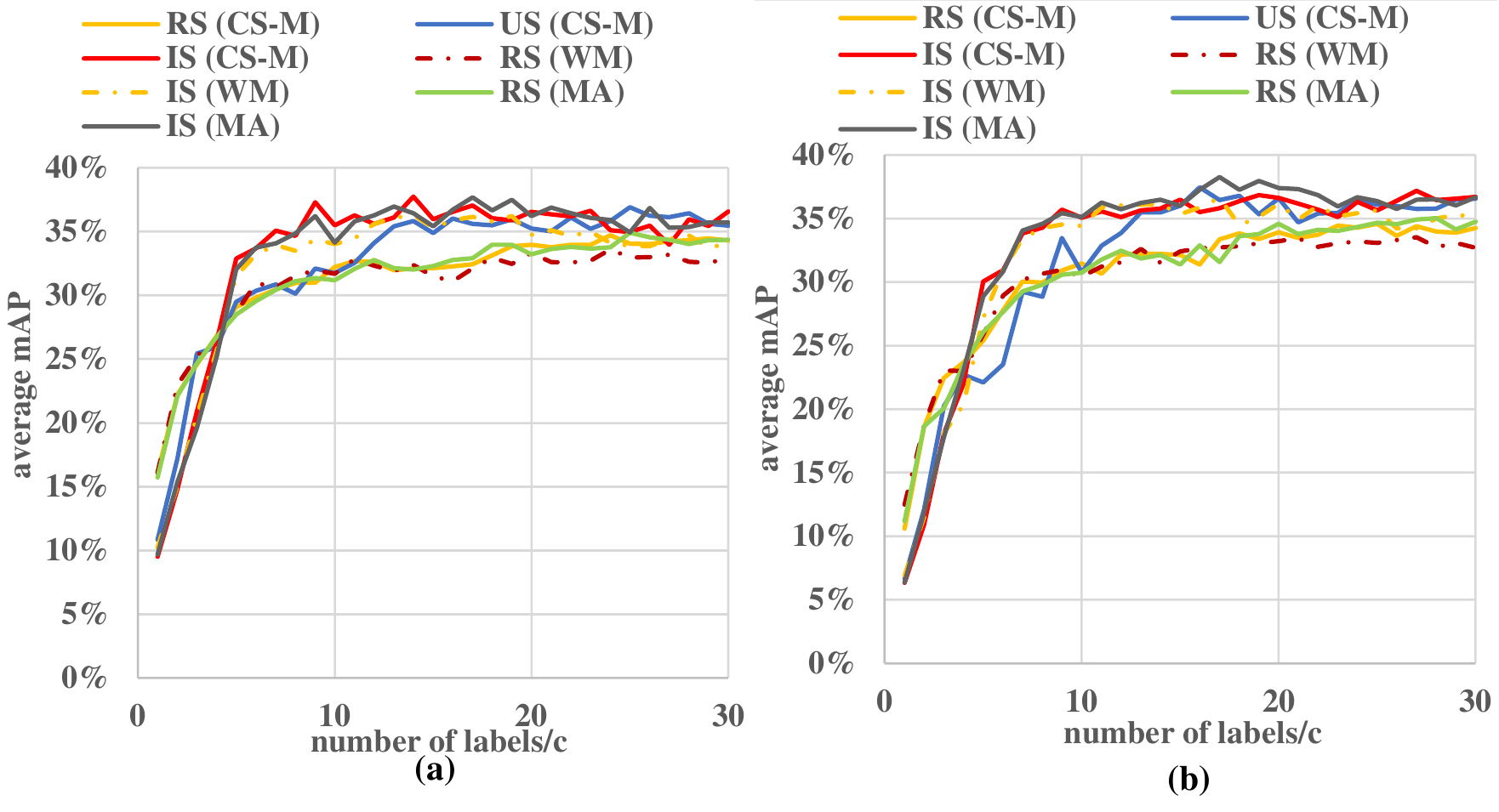}
\setlength{\abovecaptionskip}{-0.6cm}
{\caption{Comparison of different sampling strategies in Situation (1) (seed=0). (a) Training from the pre-trained model. (b) Training from scratch.}
\label{fig:sampleseed0}
}
\vspace{-6pt}
\end{figure}

\begin{figure}[h]
\centering
\includegraphics[width=
\linewidth]{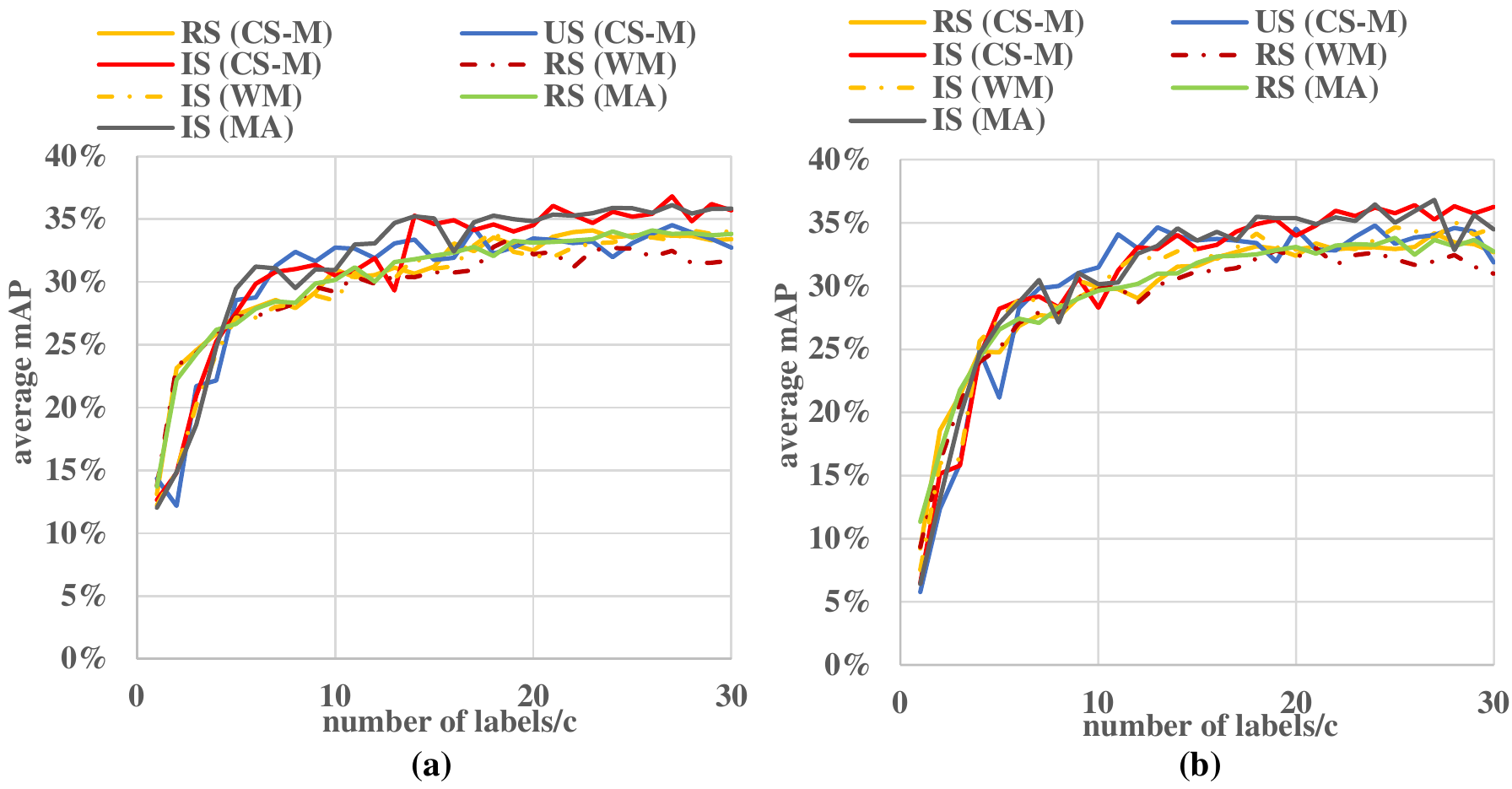}
\setlength{\abovecaptionskip}{-0.6cm}
{\caption{Comparison of different sampling strategies in Situation (1) (seed=10). (a) Training from the pre-trained model. (b) Training from scratch.}
\label{fig:sampleseed10}
}
\vspace{-6pt}
\end{figure}

\begin{figure}[h]
\centering
\includegraphics[width=
\linewidth]{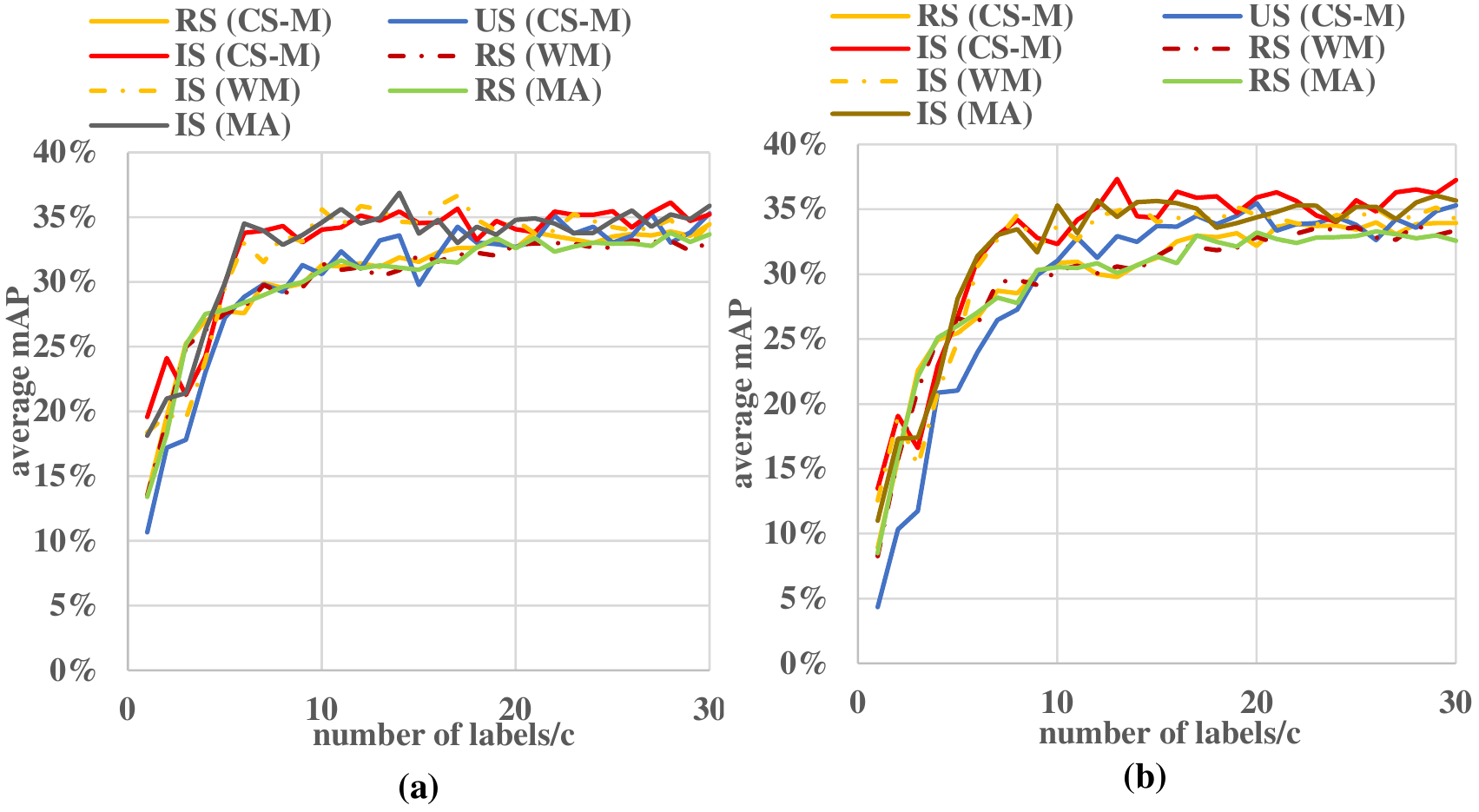}
\setlength{\abovecaptionskip}{-0.6cm}
{\caption{Comparison of different sampling strategies in Situation (2). (a) Training from the pre-trained model. (b) Training from scratch.}
\label{fig:samplecls}
}
\vspace{-6pt}
\end{figure}

\begin{figure}[h]
\centering
\includegraphics[width=
\linewidth]{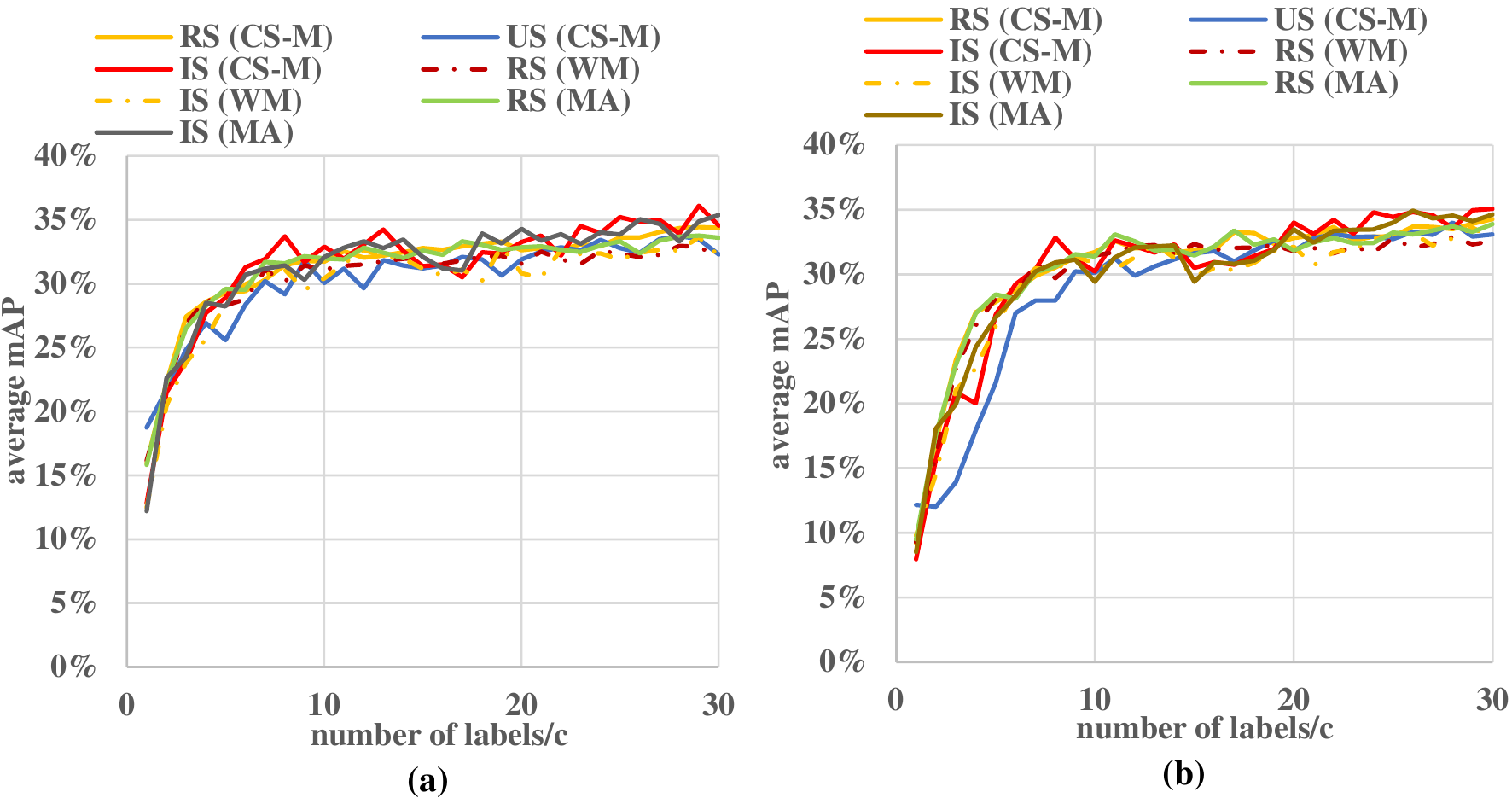}
\setlength{\abovecaptionskip}{-0.6cm}
{\caption{Comparison of different sampling strategies in Situation (3). (a) Training from the pre-trained model. (b) Training from scratch.}
\label{fig:sampleseq}
}
\vspace{-6pt}
\end{figure}

\textbf{Baselines} We compare the proposed interests-based sampling (IS) strategy with random sampling (RS) and uncertainty sampling (US). RS is the most simple sampling strategy that selects a given number of segments randomly from the unlabeled original segments. In this section, we randomly choose the original segment indexes with seed=0, seed=10, and seed=100. Then we train the TAL model with three different selected labeled segments and calculate the mean of the average mAP in each experiment. All the selected segments are labeled at the beginning of the training process. 
US selects the most uncertain samples based on the current model. In this work, we use the commonly used information entropy~\cite{yin2017deep} to measure the uncertainty. In segment $i$, we use the pre-trained model to predict the action attention value $A_{ij}$ of each clip $j$, the entropy of the whole segment is illustrated in (\ref{eq:us}). When the entropy is higher, it means that the current model is more uncertain to judge whether the clips of the segments are actions or background, so the segment should be selected to learn.

In this section, we test different combinations of merging and strategies. From Section~\ref{sec:merge_result}, we find that the performance of WM and RM is quite similar. For CS-M, the splitting approach cannot improve the learning mAPs, while increasing the merging iterations in each epoch cannot outperform CS-M (1 iter) in general. Therefore, we test WM, RM, and CS-M (1 iter) with different sampling strategies. The results of RS with CS-M, US with CS-M, IS with CS-M, RS with WM, IS with WM, RS with MA, and IS with MA are shown in Figs.~\ref{fig:sampleseed0}-\ref{fig:sampleseq}.

\textbf{Results}
As seen in Fig.~\ref{fig:sampleseed0} (a), for the proposed CS-M, in Situation (1) when the random seed=0, the proposed IS approach outperforms RS and US when the number of labeled segments is greater than 5c. The average testing mAP exceeds 36\% when sampling 9c segments, while the average testing mAPs of RS and US are around 32\%. The trend of training from scratch is similar to that of learning from the pre-trained TAL model. As seen in Fig.~\ref{fig:sampleseed10}, the IS approach outperforms RS and US when the number of labeled segments is greater than 15c if the random seed is 10. As seen in Fig.~\ref{fig:samplecls}, IS outperforms RS and US when the labeling budget is greater than 5c in Situation (2). In Situation (3), the testing mAPs of the three sampling approaches are similar, while the performance of the IS approach is slightly higher in general than other approaches when the number of labeled segments is greater than 20c. In general, compared to RS and US, the proposed IS approach can sample more representative segments to improve learning performance.

When applying the same sampling strategy, CS-M, MA, and WM achieve similar performance. For RS, the trends of CS-M and MA are close to each other, while the performance of WM is a little bit lower.

\vspace{-6pt}
\subsection{Implementation Results on Local Devices}
\label{sec:implement_result}

In this section,  to demonstrate the feasibility of the proposed stream learning approach, we analyze the communication and computation time of the proposed W-TAL stream learning on the local device. We first measure the computation time per epoch of the training process of the proposed CS-M (1 iter) approach with all segments labeled under the four combination orders mentioned in Section~\ref{sec:setup}. The training process is implemented on a GTX 1050 Ti GPU node card embedded in a desktop and on a Jetson TX2 embedded GPU, respectively. Then, we measure the memory space and time duration of both the whole video dataset and the selected representative video clips under these combination orders. The communication time can be estimated by dividing the memory space size by the transmission speed of the device. We use Gigabit Ethernet for the 1050 Ti and Wi-Fi for the TX2 to transmit the video clips to the oracle. We have tested that the upload speed of the Ethernet is around 643 Mbps and the upload speed of Wi-Fi is around 4.02 Mbps. The performance is shown in~\ref{tab:implement}.

Since the encoder is frozen, and the TAL model is only learned from the extracted features which take up only 508 MB. Therefore, the computation time of training an epoch is 16.24s on the 1050 Ti and 31.27s on the TX2 for the CoLA baseline, and it only requires around 2h on the 1050 Ti and around 4h on the TX2 to complete the whole training process. For streaming learning, since we conduct merging operations in each epoch, the whole computation time is larger than that of the baseline which only involves the training process. When the temporal correlation increases, more continuous segments share the same action labels, thus increasing the frequency of judging whether two segments need to merge. Therefore, the computation time of Situation (2) is longer than that of (1), and the computation time of Situation (3) is the longest. The whole training process costs 6-9h on the 1050 Ti and 18-30h on the TX2, which is feasible for learning on the device. As for data communication, the selected representative video clips only take up around 51 MB of space, while the whole video stream costs 15.3 GB. The Gigabit Ethernet can transmit the whole stream in $15.3~GB/643 ~Mbps = 3.2mins$, while transmitting only the selected clips costs only $51.4~MB/643~Mbps = 0.6s$. For the Wi-Fi module, transmitting the whole stream costs $15.3~GB/4.02 ~Mbps = 8.7h$, while transmitting only the selected clips costs only $51.4~MB/4.02~Mbps = 102.3s$. The main difference is in the labeling cost. For the baseline, it requires human effort to divide an 11.6-hour video into suitable segments, which is time-consuming and expensive. In our workflow, the total time duration of all the selected video clips is only 886.7s, which makes the oracle feasible and efficient to provide labels.

\begin{table}[!t]
\setlength\tabcolsep{3pt}
{\caption{Implementation performance on the local device  \label{tab:implement}
}
}
\centering
\begin{tabular}{|c|c|c|c|c|c|}
\hline
   & (1) seed=0 &(1) seed=10 & (2)
&  (3)
&  CoLA~\cite{zhang2021cola}

 \\
\hline 
 \makecell[c]{Computation \\time  on 1050 Ti\\(s/epoch)}
     &43.98
        &46.77 &51.11  &61.58  &16.24 

\\   \hline
\makecell[c]{Computation \\time  on TX2\\(s/epoch)}
     &134.92
        &137.61 &161.07  &209.04  &31.27 

\\   \hline
Video size & 51.4 MB &	51.3 MB  & 51.5 MB & 51.3 MB  & 15.3 GB	
\\

\hline
Time duration	& 886.7s &	886.7s  &	886.7s  &	886.7s  &	11.6h 
\\
\hline

\end{tabular}
\vspace{-6pt}
\end{table}

%% file: 7_conclusion.tex
\section{Conclusion}
This paper is the first attempt to directly learn from the on-device, long video stream without laborious manual video splitting for the W-TAL
task. We proposed a self-adaptive video dividing approach with a contrast score-based segment merging approach to convert the video stream into multiple segments, and the streaming learning performance is comparable with the W-TAL baseline. To request fewer labels if the cloud oracle cannot provide labels for all original segments, we explore different sampling strategies to sample the most representative segments to learn. In general, the proposed IS approach achieves higher learning performance than the RS and US approaches when the number of labeled segments is greater than 20c.